%% file: anonymous-submission-latex-2024.tex
\documentclass[letterpaper]{article} 
\usepackage{aaai24}  
\usepackage{times}  
\usepackage{helvet}  
\usepackage{courier}  
\usepackage[hyphens]{url}  
\usepackage{graphicx} 
\usepackage{natbib}  
\usepackage{caption} 
\usepackage{algorithm}
\usepackage{algorithmic}

\usepackage{amsmath}
\usepackage{amssymb}
\usepackage{svg}
\usepackage{multirow}
\usepackage{amsfonts}
\usepackage{color}
\usepackage{bbding}
\usepackage{array}

\newcolumntype{M}{>{\small\ensuremath}c}

\graphicspath{{imgs/}}

\usepackage{newfloat}
\usepackage{listings}
\DeclareCaptionStyle{ruled}{labelfont=normalfont,labelsep=colon,strut=off} 
\floatstyle{ruled}
\newfloat{listing}{tb}{lst}{}
\floatname{listing}{Listing}
\newlength\savewidth\newcommand\shline{\noalign{\global\savewidth\arrayrulewidth
		\global\arrayrulewidth 1pt}\hline\noalign{\global\arrayrulewidth\savewidth}}
\def\@fnsymbol#1{\ensuremath{\ifcase#1\or *\or \dagger\or \ddagger\or
        \mathsection\or \mathparagraph\or \|\or **\or \dagger\dagger
        \or \ddagger\ddagger \else\@ctrerr\fi}}

\title{Towards Efficient Training with Negative Samples in Visual Tracking}
\author{
    Qingmao Wei,
    Bi Zeng,
    Guotian Zeng
}
\affiliations{
    Guangdong University of Technology\\
    tsingmoe@gmail.com
}

\usepackage{bibentry}

\begin{document}

\maketitle

\begin{abstract}
    Current state-of-the-art (SOTA) methods in visual object tracking often require extensive computational resources and vast amounts of training data, leading to a risk of overfitting. This study introduces a more efficient training strategy to mitigate overfitting and reduce computational requirements. We balance the training process with a mix of negative and positive samples from the outset, named as Joint learning with Negative samples (JN). Negative samples refer to scenarios where the object from the template is not present in the search region, which helps to prevent the model from simply memorizing the target, and instead encourages it to use the template for object location.  To handle the negative samples effectively, we adopt a distribution-based head, which modeling the bounding box as distribution of distances to express uncertainty about the target's location in the presence of negative samples, offering an efficient way to manage the mixed sample training.  Furthermore, our approach introduces a target-indicating token. It encapsulates the target's precise location within the template image. This method provides exact boundary details with negligible computational cost but improving performance.  Our model, JN-256, exhibits superior performance on challenging benchmarks, achieving 75.8\% AO on GOT-10k and 84.1\% AUC on TrackingNet. Notably, JN-256 outperforms previous SOTA trackers that utilize larger models and higher input resolutions, even though it is trained with only half the number of data sampled used in those works.
\end{abstract}

\begin{figure}[t]
    \centering
    \includegraphics[width=0.97\columnwidth]{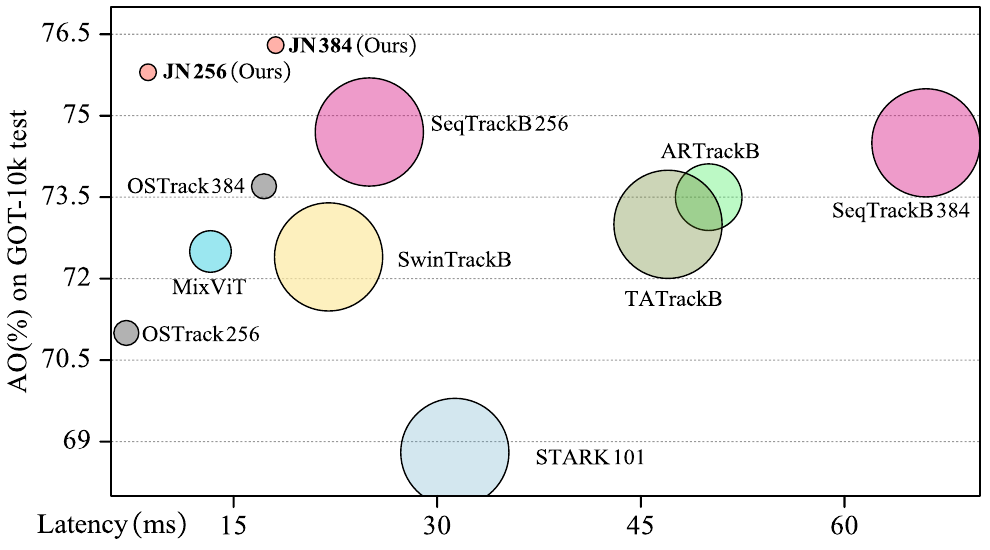} 
    \caption{Comparison of accuracy vs. latency trade-off for different visual tracking methods, with bubble size representing the training epochs.  Trained with the train split of GOT-10k, our tracker JN-256 achieve a amazing 75.8\% AO, showing impressive one-shot tracking performance.
    }
    \label{title}
\end{figure}
\section{Introduction}
\begin{figure*}[]
    \centering
    \includegraphics[width=0.97\linewidth]{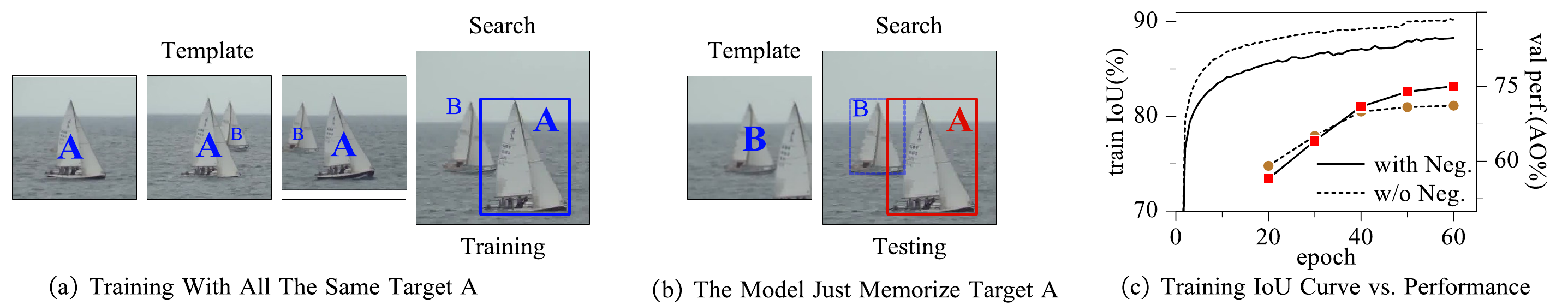} 
    \caption{When the tracker is consistently presented with different frames of the same template as in (a), it may overfit by simply memorizing the target instead of learning to locate it based on the template, as shown in (b). This issue can be alleviated by introducing negative pairs, where the target does not appear in the search region. As demonstrated in (c), training with negative samples aids the model in achieving a higher average overlap, even though its training Intersection over Union (IoU) may be lower than that achieved with all-positive training.}
    \label{illu}
\end{figure*}

Visual object tracking represents a significant challenge in computer vision and has applications in diverse domains such as surveillance, robotics, and autonomous driving. The recent integration of the Transformer~\cite{vanilla_selfattention} in object trackers~\cite{transt,TMT,ostrack,mixformer,CSWinTT,simTrack,aiatrack} has added both promise and complexity to the field. While deep learning advancements have led to notable progress, training state-of-the-art (SOTA) object trackers with Transformers remains daunting, requiring substantial time and computational resources.

STARK~\cite{STARK} and TATrack~\cite{TATrack}, who interpret the Transformer as a standalone module for feature fusion and train it from scratch, requiring an extensive amount of training samples, a problem partially mitigated by the fast convergence of works like OSTrack~\cite{ostrack} and MixFormer~\cite{mixformer}, which utilize mask-image-modeling pretraining~\cite{mae,cae,simMIM}.

However, we identify two main areas for further optimization and propose novel solutions to address them:

First, current visual tracking frameworks typically use an image pair of a template and a search region. The template image, though representing the target, offers only a vague hint of its location. To provide the model with more precise information about the target's position, we introduce a Target-Indicating Token (TIT). This token is a carefully crafted input that encodes specific details about the target's exact location within the template image. By feeding the TIT into the Transformer along with the image features, the model receives a direct cue to the target's precise boundaries. Unlike previous methods~\cite{ostrack,mixformer}, which may have only indirectly inferred the target's position, the inclusion of TIT equips the model with direct and accurate positional data about the target object. This contributes to significantly more accurate tracking performance by providing an additional layer of information.

Second, previous works often solely rely on positive pairs in training, potentially leading to overfitting. To combat this, we introduce Joint Learning with Negative samples (JN), utilizing a balanced mix of negative and positive samples from the outset. This method goes beyond the typical training routine. By incorporating negative samples—instances where the target is absent from the search region—the model learns to develop an awareness of the possibility of the target's absence. This awareness forces the model to rely more heavily on the template when searching for the target in the search region, rather than simply memorizing the target. We further support this mechanism by adopting a distribution-based head from the Generalized Focal Loss~\cite{GFL, GFLv2}. This head predicts a distribution of distances, expressing uncertainty about the target's location when negative samples are present. This nuanced approach enables more effective learning from mixed samples, leading to improved object tracking performance..

Despite training with fewer samples due to the mix of negative image pairs, our method achieves impressive performance, setting a new SOTA on multiple benchmarks. It maintains admirable inference efficiency and shows faster convergence compared to SOTA Transformer-based trackers. This method achieves a balance between accuracy and inference speed, as demonstrated in Fig.~\ref{title}.

In summary, our work contributes in the following ways:

\begin{itemize}
\item We introduce a novel approach of simultaneous learning of classification and localization with negative samples, significantly enhancing visual object tracking performance, especially with the distribution-based head.
\item We propose a Target-Indicating Token that efficiently provides location annotations for the target within the template without significant computational cost.
\item Our experiments show that our model, even with reduced training samples, outperforms existing state-of-the-art trackers on challenging benchmarks.
\end{itemize}

\section{Related Work}

\begin{figure*}[t]
    \centering
    \includegraphics[width=0.9\textwidth]{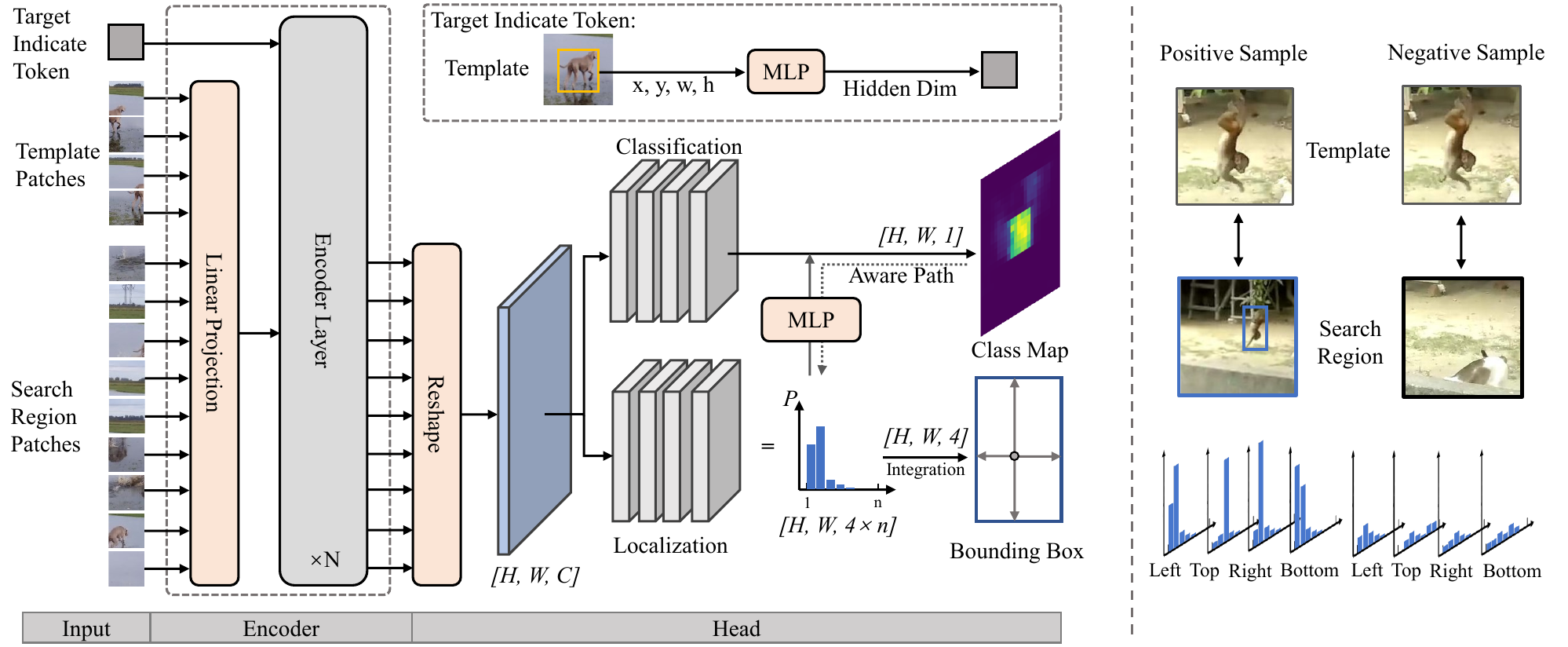} 
    \caption{Illustration of our proposed pipeline. The template and search region are split, flattened, and projected then concatenated with a Target-Indicating Token.  For the positive image pairs, the supervision flows on two branches. For the negative image pairs, the localization branch is supervised by the loss from the \textit{Aware Path}.}
    \label{fig_arch}
\end{figure*}

\textbf{Training Pipeline in Object Tracking.}
The training pipelines of previous trackers are often complex. STARK~\cite{STARK} requires numerous positive samples for localization training in its initial stage, followed by a second stage that trains a separate classification module for online tracking. Mixformer~\cite{mixformer} follows a similar training paradigm, even though it employs the Transformer for both feature extraction and relation modeling. The classification module in Mixformer is more efficient than STARK, but it still requires a split stage for training. In contrast, as an offline tracker, OSTrack~\cite{ostrack} solely trains the model with positive samples throughout the entire training process. However, similar to the first training stage of online trackers, the localization branch in OSTrack predominantly encounters positive samples, which may lead to overfitting issues.
UniTrack~\cite{UniTrack}, Unicorn~\cite{unicorn}, and UNINEXT~\cite{UNINEXT} leverage both single object tracking (SOT) and multi-object tracking (MOT) datasets, allowing their models to handle multiple tasks such as SOT, MOT and video object segmentation within a unified framework. However, these methods yield only ordinary performance on the one-shot setting benchmark, GOT-10k~\cite{GOT-10k}, which specifically requires trackers to be trained solely on its own train split. This suggests that these methods heavily rely on extra data to achieve performance improvement.
To encourage the tracker to utilize information from the template instead of merely memorizing the target in the search region during the training process, we introduce negative image pairs throughout the entire training stage. 

\textbf{Multi-modality of Transformer.}
Traditionally developed for natural language processing, the Transformer architecture has demonstrated its adaptability to diverse data modalities. ViT~\cite{vit} concatenate a learnable \texttt{CLS} token to the input image embeddings to extract category information, demonstrating the potential of transformers in modeling image features along with other modalities. SwinTrack~\cite{SwinTrack} attempts to introduce the Transformer Decoder for feature interaction in object tracking and achieves improved performance by incorporating the Motion Token, which contains historical position information of the target. This further proves the transformer's capability to handle different modalities of information. Leveraging recent developments in one-stage object tracking, we design a target-indicating token that, like ViT, directly concatenates with the input image patch. This token explicitly informs the model about the precise position of the target in the template image.

\section{Method}

\subsection{Model Overview}

The overall design of our model is presented in Fig.~\ref{fig_arch}. It adopts a simple encoder-head architecture following recent tracking works~\cite{ostrack,mixformer}.

\textbf{Encoder.}
The encoder can be a general image encoder that encodes pixels into hidden feature representations, such as ConvNet, vision Transformer (ViT), ora hybrid architecture. In this work, we use the same ViT encoder as OSTrack~\cite{ostrack} for visual feature encoding. The template and search images are first split into patches, flattened and projected to generate a sequence of tokens and added with positional embedding. The tokens are concatenated with an extra target-indicating token which carries the location of the target in the template image as detailed in Sec.~\ref{sec:target_token}. we feed them into a plain ViT backbone to encode visual features. We provide more details and discussion about the encoder in Appendix A.

\textbf{Head.}
The goal of the head is to predict the bounding box or the mask of the target. In our visual object tracking model, we primarily consider three types of head modules: Center Head, Corner Head, and Distribution-based Head.

The Center Head, utilized by OSTrack, comprises a lightweight, fully convolutional module. It consists of three branches: a centerness branch, an offset branch, and a size branch. The centerness branch classifies grid cells based on how close they are to the object's center, while the offset branch refines this by providing an offset to the exact center from the predicted grid. The size branch regresses the width and height of the object. The outputs of these branches together yield a precise bounding box.

The Corner Head, employed by Mixformer~\cite{mixformer} and STARK~\cite{STARK}, operates by predicting the top-left and bottom-right corners of the bounding box. Each corner prediction is treated as a classification task with its own specific branch, which aims to accurately place the corners by generating expected values in two dimensions.

The Distribution-based Head, inspired by the Generalized Focal Loss~\cite{GFLv2} in object detection, consists of a quality-aware classification branch and a distribution-based localization branch. As shown in the head part of Fig.~\ref{fig_arch}, the localization branch models the output box as the distribution-based of left, top, right, and bottom (ltrb) distances. 
Given a discretizing resolution $n$, it predicts the probabilitis of distance $y$ with minimum $y_0$ and maximum $y_n (y_0 \leq y \leq y_n, n \in \mathbf{N}) $. The estimated distance $\hat{y}$ can be integrated from the the output:
\begin{equation}
\hat{y}=\int_{y_0}^{y_n} P(x) x \mathrm{~d} x =\sum_{i=0}^n P\left(y_i\right) y_i
\end{equation}
We provide more details about setting the discretizing resolution $n$ in Appendix C.

With prediction of distribution of distance on each direction, top-k probabilitis of distribution and the mean value of them are sampled as $\operatorname{topkm(\mathbf{P}^\omega)}$ in each direction $w$ output and then concatenated as a statistical feature $\mathbf{F} \in \mathbb{R}^{4(k+1)}$, which stands for the certainty of the distribution:
\begin{equation}
\mathbf{F} = \operatorname{concat}(\{\operatorname{topkm(\mathbf{P}^\omega) | \omega \in \{l,t,r,b\}}\})
\end{equation}
The final output of quality-aware classification score is presented as $\sigma$:
\begin{equation}
    \sigma = \mathbf{C} \cdot \mathcal{M}(\mathbf{F})
\end{equation}
where $\mathcal{M}$ is a simple multi-layer perceptron(MLP) to project the statistical feature $\mathbf{F}$ with the dimension of original classification score $\mathbf{C}$. $\mathcal{M}$ serverd as a \textit{Aware Path} between the branches, passing the supervision signal from the classification to the localization branch.  The process is shown as head part in Fig.~\ref{fig_arch}.

While we found that the distribution-based head performs best in our model in subsequent experiments, this doesn't mean the other two head modules are without merit. In Sec.~\ref{loss}, we will delve deeper into how these three head modules behave when dealing with negative samples, and their loss function design.

\begin{figure}[t]
    \centering
    \includegraphics[width=0.97\linewidth]{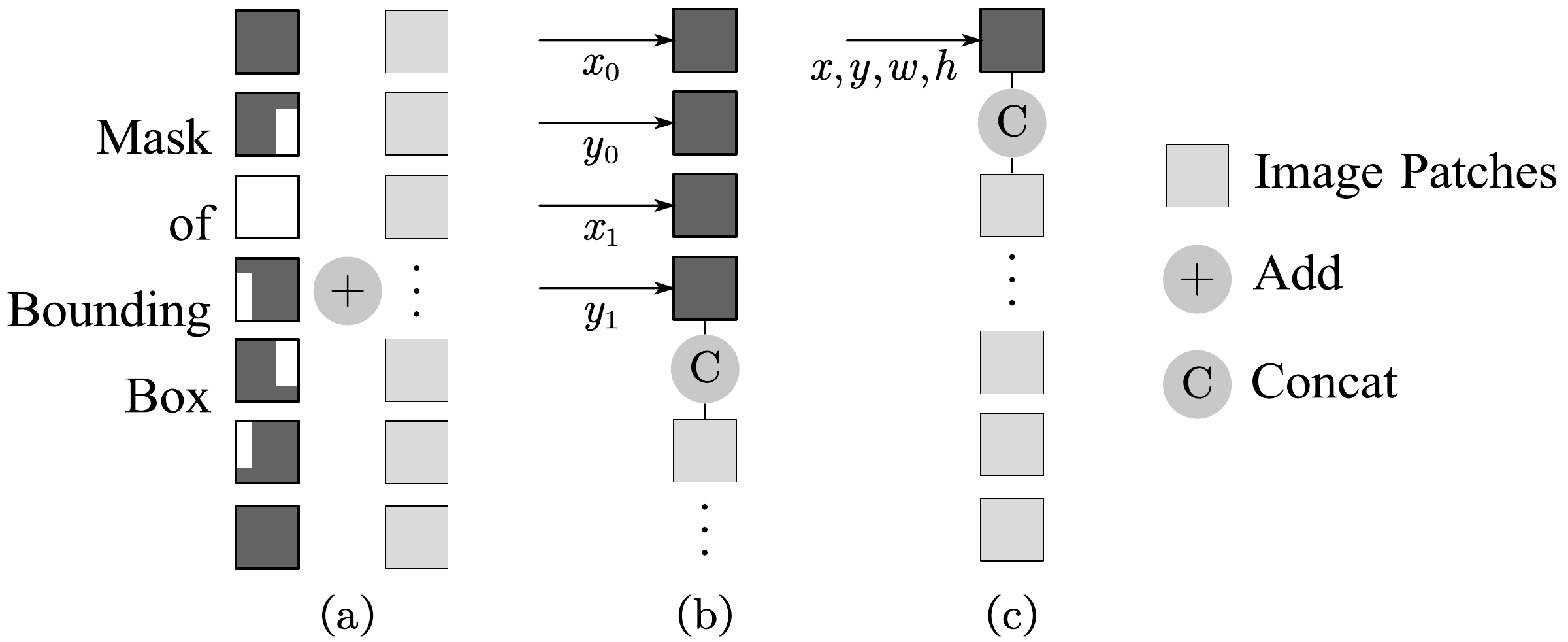} 
    \caption{Comparisons of ways of position inception.  The projection modules e.g. conv or MLP are omitted. (a): adding the embedding to the features; (b) \& (c): concatenating the tokens to the features. }
    \label{inception}
\end{figure}

\subsection{Target-Indicating Token}
\label{sec:target_token}
In the context of object tracking under image-pair settings, the template image is a cropped image centered around the target to be tracked.  Throughout both the training and test phases, the precise position of the target is known, typically represented as a bounding box or a mask.  However, many recent trackers fail to fully exploit this valuable position information.

To address this limitation, some trackers, e.g. STMTrack~\cite{stmtrack} and attempt to incorporate the position information by adding an embedding genrated from the pseudo mask of the target to the extracted template features as shown in Fig.~\ref{inception}(a).  However, we observed that this approach can introduce noise and adversely affect the feature representation.
We propose a novel approach - concatenating the embedding tokens instead of adding it to the features.  The tokens, denoted as Target-indicating Token, is formulated to represent the bounding box information and are concatenated with the image features fed into the encoder.  Given the coordinates of the box of the target, a simple way is to use four tokens carries each number of the coordinates shown in Fig.~\ref{inception}(b).   In our method, we leverage an multi-layer perceptron $\mathcal{M}$ to project the box information in one single Target-indicating Token (TIT) depicted in Fig.~\ref{inception}(c), as follows:
\begin{equation}
\boldsymbol{H}_{i} = \mathbf{\mathcal{M}}(x_z, y_z, w_z, h_z)
\end{equation}
where $(x_z, y_z, w_z, h_z)$ is the bounding box of the target in the template.  As illustrated in Fig.~\ref{fig_arch}, TIT is concatenated with the visual feature tokens as the input of the Transformer Encoder.   This innovative design enables our tracker to gain a deeper understanding of the target's spatial location and directly incorporate this critical information into the feature extraction process as interpreted in Fig.~\ref{attn_map}, while introducing negalible computation.  More discussion about how TIT interacted with the image features are provided in the Appendix A \& B.

\subsection{Joint Learning with Negative Samples}
\label{sec:negative_samples}
As the input of the model is an image pair of the template image and search region, we define that (1) \textbf{Positive} sample is a pair where the target in the template image appears in the search region, and (2) \textbf{Negative} sample is a pair where the target in the template image does not appear in the search region.  It was observed in STARK~\cite{STARK} that joint learning of localization and classification, i.e., training with both positive and negative samples within a single stage, led to sub-optimal results.  This could be attributed to the original design of the localization head, which was found to be incompatible with the classification task.  Therefore, drawing inspiration from the head design in GFL~\cite{GFL}, we introduced a head based on uncertainty estimation into the tracker.  Using this head, we explored the effectiveness of simultaneous classification and localization learning within this framework by mixing up the negative samples during training.  With discretized distribution representation of bounding box, we encourage the box head to predict an ambiguous distribution facing negetive samples.  

\textbf{Data Sampling:}
In our training pipeline, we employ a joint learning strategy that involves mixing negative image pairs with positive ones using a ratio denoted as $\rho \in (0,1]$.  This ratio determines the proportion of positve samples in the training data.  For positive image pairs, we randomly sample two frames from a sequence within the maximum sampling gap, ensuring that both frames contain the visible target.
When generating negative image pairs, we start with a randomly sampled template.  To create a negative pair, we first attempt to sample a frame from the same sequence as the template but with an absence label, indicating that the target is not present in the frame.  If the target is consistently present in the sequence of the template image, we randomly sample a frame from another sequence, contributing to a more comprehensive representation of negative scenarios.
As we do not supervise the localization in negative pairs, the effective number of samples used for training localization is less than that for training classification.  The relationship between the effective number of samples in joint learning ($n^{*}$) and that in the all-positive learning ($n'$) is given by:
\begin{equation}
\begin{split}
n^{*} = \rho \cdot n'
\end{split}
\end{equation}
In summary, the parameter $\rho$ controls the balance between positive and negative samples in the training data, and its value directly influences the effective number of samples used for training localization. Adjusting $\rho$ allows us to fine-tune the training process and optimize the model's performance in object tracking tasks, which will be detailed discussed in Sec.~\ref{ConvergenceandRatio}.

\subsection{Loss Function with Negative Samples} 
\label{loss}
As we introduce negative pairs into the training process, we encounter differences in label assignment and loss function for different heads. 

\input{got10k_table}
(1)\textit{With Center Head:}
The output box is supervised by $l1$ loss and $\operatorname{GIoU}$ loss~\cite{generalized_iou}.  Focal loss(FL)~\cite{focal} is aopted for the classification map.  For the negative pairs, we only assign the supervise the center branch by simply assign all the label to zeros.  The final loss of center head can be written as:
\begin{equation}
\begin{split}
    \mathcal{L}_{\operatorname{center}} &=
    \frac{1}{N_{pos}}(
        \lambda_{l1}L_{l1}
        + \lambda_{\operatorname{GIoU}}L_{\operatorname{GIoU}}
        + \lambda_{\operatorname{FL}}L_{\operatorname{FL}}
    )\\
    &+ 
    \frac{1}{N_{neg}} \lambda_{\operatorname{FL}}L_{\operatorname{FL}}
\end{split}
\end{equation}

(2)\textit{With Corner Head:}
For positive pairs, the ouput box is supervised by combination of $l1$ and $\operatorname{GIoU}$ following Mixformer.
For the negative pairs, no integration is performed as no object in search region, thus we directly assign both the classification maps to all zeros and apply a standard cross-entropy(CE) loss on it.  The final loss of corner head can be written as:
\begin{equation}
    \begin{split}
        \mathcal{L}_{\operatorname{corner}} &=
        \frac{1}{N_{pos}}(
            \lambda_{l1}L_{l1}
            + \lambda_{\operatorname{GIoU}}L_{\operatorname{GIoU}}
        )\\
        &+ 
        \frac{1}{N_{neg}} \lambda_{\operatorname{CE}}L_{\operatorname{CE}}
\end{split}
\end{equation}

(3)\textit{With Distribution-based Head:}
For positive image pairs, the classification branch is encourage to predict the quality of localization, e.g. the intersection over union(IoU) between the predicted box and the Ground Truth (G.T.) box.  Therefore, we assign the calculated IoU to the points inside, while the points outside the G.T. box are assigned to zeros during training.  The target of classification is supervised by quality focal loss(QFL)~\cite{GFL}:
\begin{equation}
    \mathcal{L}_{\operatorname{QFL}}(\sigma)=-|y-\sigma|^\beta((1-y) \log (1-\sigma)+y \log (\sigma))
\end{equation}
Unlike corner head, which only calculates the loss on the expected box for localization, here we directly supervise the distribution of disctance with Distribution Focal Loss(DFL):
\begin{equation}
    \mathcal{L}_{\operatorname{DFL}} (\mathcal{S}_i, \mathcal{S}_{i+1}) = -((y_{i+1}-y)\mathrm{log} (\mathcal{S}_i)
+(y-y_i)\mathrm{log} (\mathcal{S}_{i+1}))
\end{equation}
The global minimum solution of DFL comes when $\mathcal{S}_i = \frac{y_{i+1}-y}{y_{i+1}-y_i}, \mathcal{S}_{i+1} = \frac{y-y_i}{y_{i+1}-y_i}$, thus focus on enlarging the probabilities of the values around target $y$.
For negative image pairs, we consider all points from the feature map as negative examples, setting the class label of all these points uniformly to zero.  This approach ensures an appropriate distribution of positive and negative samples, facilitating effective training and accurate predictions in our model.  The final loss of distribution-based head can be written as:
\begin{equation}
    \begin{split}
        \mathcal{L}_{\operatorname{dist.}} &=
        \frac{1}{N_{pos}}(
            \lambda_{l1}L_{l1}
            + \lambda_{\operatorname{GIoU}}L_{\operatorname{GIoU}}
            + \lambda_{\operatorname{DFL}}L_{\operatorname{DFL}}\\
            &+ L_{\operatorname{QFL}}
        ) + 
        \frac{1}{N_{neg}} \lambda_{\operatorname{QFL}}L_{\operatorname{QFL}}
\end{split}
\end{equation}

We set $\lambda_{l1} = 5$ and $\lambda_{GIoU} = 2$ in all the above heads as in~\cite{ostrack}.  In distribution-based head, we set $\lambda_{\operatorname{DFL}} = 0.2$ as in~\cite{GFL}.

\input{head_table}

\section{Experiments}
\input{lasot_tknet_table}
\input{vot2020_table}

\subsection{Setup}
\label{setup}
\textbf{Implementation Details:}
The backbone we use is a standard transformer encoder from ViT-Base.  We initialize the weight of encoder from pretrained parameters by CAE~\cite{cae}.
For testing the efficiency of our method to the baseline, we proposed two versions with different input resolutions:
\begin{itemize}
\item \textbf{JN-256.}  Backbone: ViT-Base; Search region: 256$\times$256 pixels; Template: 144$\times$144 pixels.
\item \textbf{JN-384.}  Backbone: ViT-Base; Search region: 384$\times$384 pixels; Template: 192$\times$192 pixels.
\end{itemize} 

\textbf{Training.}
For the GOT-10k~\cite{GOT-10k} benchmark, we only use the training split of GOT-10k following the one-shot protocols and train the model for 60 epochs.  For the other benchmarks, the training splits of GOT-10k, COCO~\cite{COCO}, LaSOT~\cite{LaSOT} and TrackingNet~\cite{TrackingNet} are used for training in 180 epochs. Common data augmentations including horizontal flip and brightness jittering are used in training.  For JN-256, we train the model on a single RTX 3090 GPU, holding 64 image pairs in a batch.  For JN-384, we train the model on two RTX 3090 GPUs, holding 28 image pairs per GPU, resulting in a batch size of 56.  We train the model with adam~\cite{adamw} optimizer, set the weight decay to $10^{-4}$. The initial learning rate of backbone for both 256 and 384 version are set as to $2\times10^{-5}$ and other paremeters to $2\times10^{-5}$.  We decrease the learning rate by the factor of 0.1 in the last 10\% epochs.  We set defalut ratio of positive samples $\rho$ to 0.7.

\textbf{Inference.} During inference, we adopt Hanning window penalty to utilize positional prior like scale change and motion smoothness in tracking, following the common practice~\cite{SiamFC, SwinTrack, ostrack}. The output score map is simply element-wise multiplied by the Hanning window with the same size, and we choose the box with the highest multiplied score as the target box.

\subsection{Results and Comparisons}

\textbf{GOT-10k.}
GOT-10k~\cite{GOT-10k} is a large-scale generic object tracking benchmark with 10000 video sequences, which includes 180 videos for testing. Note that it is zero-class overlap between the train subset and test subset, which requires the ability to generalization of a tracker. Following the one-shot protocol, we train our models only with the GOT-10K train set and evaluate it with other state-of-the-art tracking methods on the test set.  As reported in Tab.~\ref{table:got10k}, with the same ViT-Base backbone, our base model JN-256 achieve amazingly 75.8\% Average Overlap(AO). Note that the our model is trained with only 60 epoch, and keeps excellent speed-accuracy trade-off.  Furthermore, JN-384 obtains a new state-of-the-art AO score of 76.3\%, surpassing ARTrack-384 by 0.9\%. The results prove the generalization ability of our proposed method on unseen target classes.

\textbf{TrackaingNet.}
TrackingNet~\cite{TrackingNet} is a large-scale short-term tracking benchmark containing 511 video sequences in the test set, which covers diverse target classes. As reported in Tab.~\ref{table:tknet_lasot}, our JN-384 achieve 84.9 \%AUC, surpassing OSTrack-384 by 1.0\%. Notice that we trained our model only with half image pairs of that in OSTrack.  Moreover, our JN-256 with input resolution of 256$\times$256 also performs better than OSTrack-384 with input resolution of 384$\times$384.

\textbf{VOT2020.}
VOT2020~\cite{VOT2020} is a challenging short-term tracking benchmark that is evaluated by target segmentation results. To evaluate JN, we use AlphaRefine~\cite{Alpha-Refine} to generate segmentation masks, and the results are shown in Tab.~\ref{table:vot2020}. The overall performance is ranked by the Expected Average Overlap (EAO). Our tracker exhibits very competitive performance even with a 256-input base model.

\textbf{LaSOT.}
LaSOT~\cite{LaSOT} is a challenging long-term tracking benchmark, which contains 280 videos for testing. We compare the result of our JN with SOTA trackers in Tab.~\ref{table:tknet_lasot}, JN-256 achieves an AUC of 67.9\%, while JN-384 slightly improves the score to 68.5\% AUC. Despite the less pronounced difference, these results are substantial, considering the complexities of long-term tracking.

\subsection{Ablation Study}

\textbf{Joint Learning with Different Heads.}
We examined the joint learning strategy (comprising 70\% positive and 30\% negative samples) on different head designs: center, corner, and distribution-based heads. As shown in Tab.~\ref{integration}, when applying joint learning, the performance of the center head slightly declines. This might be due to the limited interaction between its branches. Surprisingly, the corner head shows an improvement with joint learning, which is different from the findings in STARK where joint learning negatively affected the corner head. One potential reason for this difference could be the sample ratio used in STARK's joint learning. In our method, the presence of the \textit{Aware Path} between the localization and classification branches seems to enhance the benefits of joint learning. This setup allows more interaction than the center head without causing the issues observed in the corner head.

\textbf{Different Ways of Inception of Target Position.}
\input{inception_table}
Except for our proposed target-indicating token, we test different methods (as depicted in Fig.~\ref{inception}) for the integration of target position information, as shown in Tab.~\ref{table:inception}. By adding the embedding of a pseudo mask generated from the bounding box, we observe a decrease in performance compared to the baseline without any integration. Concatenating the tokens enhances the performance, but interestingly, using only one token appears to be slightly superior to employing four tokens for each number. Additionally, we experiment with sine-cosine position embedding, which yields some improvement in performance compared to the baseline. However, this improvement is not as significant as when using a Multilayer Perceptron (MLP).

\textbf{Convergence and Ratio of Positive.}
\label{ConvergenceandRatio}
\begin{figure}[t]
    \centering
    \includegraphics[width=0.97\linewidth]{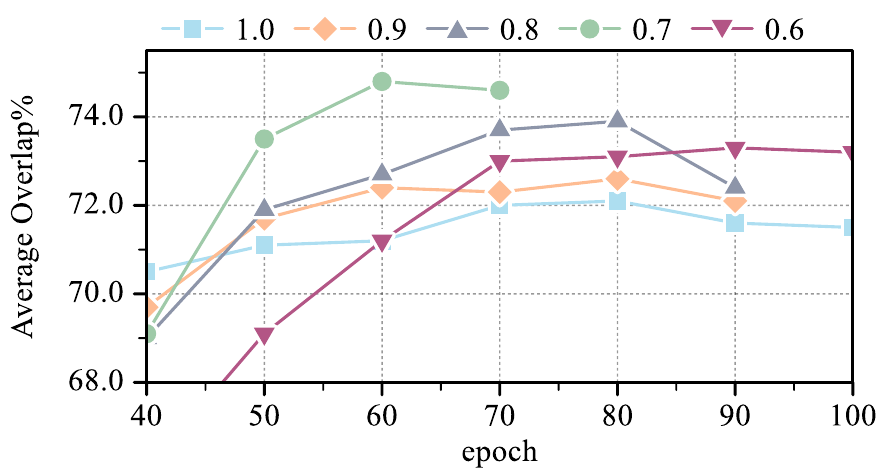} 
    \caption{Average Overlap on GOT-10k validation set with different ratio $\rho$ and different epochs.}
    \label{ratio}
\end{figure}
We perform experiments on the GOT-10k validation set to discern our tracker's convergence time and the best negative sample ratio for a balance between accuracy and efficiency. For stability, we compute the mean Average Overlap (AO) from the final five epochs to lessen performance fluctuations. The experimental procedure is detailed in Sec.~\ref{setup}, with the learning rate reduced to 10\% during the final 20\% of epochs.
Fig.~\ref{ratio} indicates our tracker's convergence within 60 epochs on GOT-10k, reaching an AO of 74.3\%. While training exclusively on positive samples gives better AO for fewer epochs, introducing negative samples enhances performance over extended training. Yet, overloading with negative samples can degrade results. We find the optimal performance at a ratio, $\rho=0.7$, emphasizing the significance of an appropriate negative sample ratio in training.

\begin{figure}[t]
    \centering
    \includegraphics[width=0.97\linewidth]{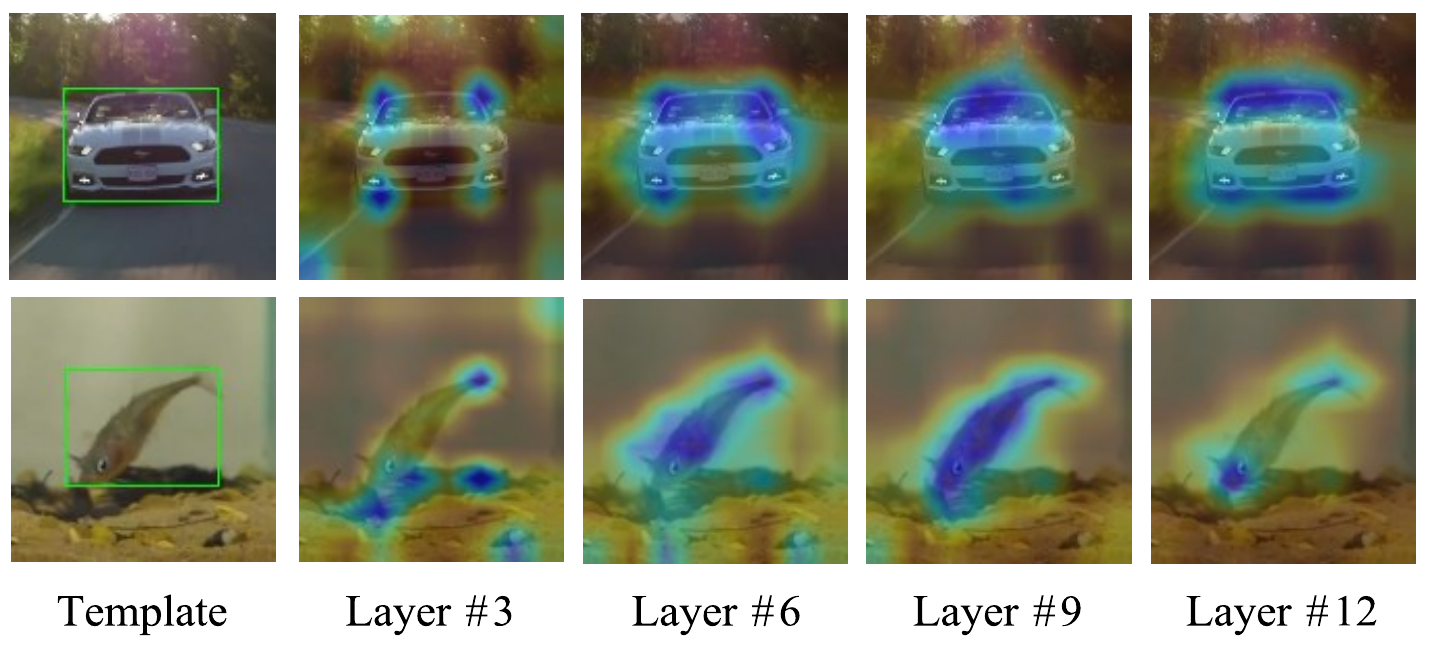} 
    \caption{Visualization of attention map (of TIT attending to the tokens from the template) in the encoder.}
    \label{attn_map}
\end{figure}

\begin{figure}[t]
    \centering
    \includegraphics[width=0.97\linewidth]{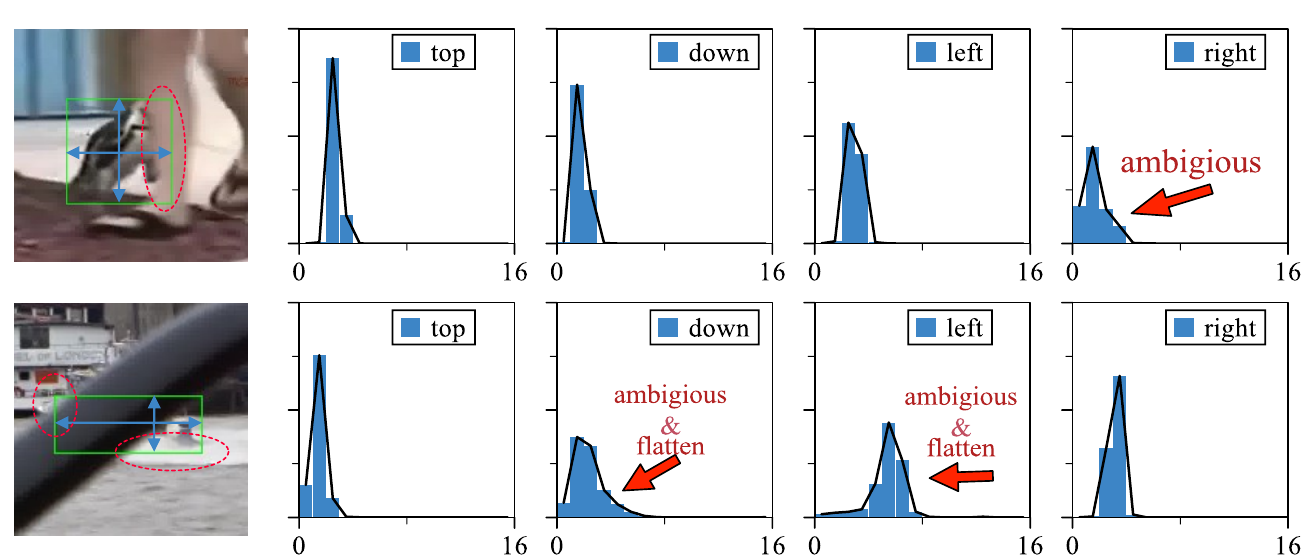} 
    \caption{Due to occlusion, the boundaries of objects are not clear enough, thus the model predicts uncertain distribution.  X-axis in each distribution diagram denotes for the non-normalized distance to the start point.}
    \label{dist}
\end{figure}

\textbf{Visualization.}
In Fig.~\ref{attn_map}, we visualize the attention from the TIT to other image tokens in the template by averaging attention weights across heads, reshaping to 2D, and upscaling to the image's resolution. By visualization, the utility of TIT in aiding target recognition becomes evident.
Moreover, Fig.~\ref{dist} depicts the distribution of distances from the point with the highest classification score in the distribution-based head. This highlights the model's response to partial occlusions, such as a penguin's head obscured by feet or a boat mostly hidden behind a fence. In these cases, the model presents a spread-out, ambiguous distance distribution.  More visualization are provided in Appendix F.

\section{Conclusion}
In this work, we introduce JN, a tracker that incorporates joint learning with negative samples. Extensive experiments demonstrate the effectiveness of our joint learning strategy, proving its utility in conjunction with a distribution-based head adopted from existing designs. Furthermore, we propose a target-indicating token that enhances the tracker's utilization of the template by directly providing the target's location within the template image, demonstrating the multi-modality of the Transformer. Even with fewer training samples, our method achieves superior performance on shor-term tracking benchmarks. We hope this work will inspire further research on training efficiency in object tracking.

\input{supplementary.tex}

\bibliography{aaai24,vot}
\end{document}

%% file: got10k_table.tex
\begin{table*}[!ht]
    \centering
    \small
    \begin{tabular}{c|c|c|c|ccc|c|c}
        \shline
        \rule{0pt}{9pt}
        Method                        & Backbone   & Input Size & Epochs & AO(\%)                  & SR$_{0.50}$(\%)          & SR$_{0.75}$(\%)        & Params   & FPS   \\
        \shline
        JN(Ours)                      & ViT-Base   & 256      & 60   & \underline{75.8}  & \textbf{86.3}    & 72.9 & 68.7M  & 114 \\
        JN(Ours)                      & ViT-Base   & 384      & 60   & \textbf{76.3}     & \underline{84.9} & \textbf{76.1}  & 68.7M  & 55  \\
        \hline
        ARTrack~\cite{ARTrack}        & ViT-Base   & 384      & 120  & 75.5              & 84.3             & \underline{74.3}           & 173.1M & 23  \\
        SeqTrack\cite{SeqTrack}       & ViT-Base   & 384      & 500  & 74.5              & 84.3             & 71.4           & 89M    & 15  \\
        GRM\cite{GRM}            & ViT-Base   & 256      & 100  & 73.4              & 80.4             & 68.2           & 99.4M  & 79  \\
        OSTrack\cite{ostrack}         & ViT-Base   & 384      & 100  & 73.7              & 83.2             & 70.8           & 70.2M  & 58  \\
        Mixformer\cite{mixformer}     & MixViT     & 288      & 180  & 72.5              & 82.4             & 69.9           & 97M    & 75  \\
        
        SimTrack\cite{simTrack}       & ViT-Base   & 256      & 500  & 68.6              & 78.9             & 62.4           & -        & 94$^*$  \\
        AiATrack\cite{aiatrack}       & ResNet-50  & 256      & 100  & 69.6              & 80.0             & 63.2           & -        & 38$^*$  \\

        SwinTrack\cite{SwinTrack}     & Swin-Base  & 384      & 300  & 72.4              & 80.5             & 67.8           & 91M    & 45  \\
        TATrack\cite{TATrack}         & SwinB+LCA  & 224      & 500  & 73.0              & 83.3             & 68.5           & 112.8M & 21  \\
        STARK\cite{spatial-frequency} & ResNet-101 & 320      & 500  & 68.8              & 78.1             & 64.1           & 42M    & 32  \\
        TransT\cite{transt}           & ResNet-50  & 256      & -      & 67.1              & 76.8             & 60.9           & -        & -     \\
        SiamR-CNN\cite{SiamR-CNN}     & ResNet-101 & 1333     & -      & 64.9              & 72.8             & 59.7           & -        & -     \\
        Ocean\cite{Ocean}             & ResNet-50  & 255      & 50   & 61.1              & 72.1             & 47.3           & -        & -     \\
        \shline
    \end{tabular}
    \caption{Results for JN and other tracking models on GOT-10k test set trained with only train split of GOT-10k (the so called one-shot setting). The best two results are shown in \textbf{bold} and \underline{underline}. The Frame Per Second(FPS) is counted on a RTX 2080Ti GPU. * indicates that the FPS of models are from the paper tested with a more powerful GPU than 2080Ti.}
    \label{table:got10k}
\end{table*}

%% file: head_table.tex
\begin{table}[]
    \centering
    \begin{tabular}{c|c|lll}
        \shline
        \rule{0pt}{9pt} Heads                   & J.L.       & {AO(\%) }                                                        & SR$_{0.50}$(\%)                                                   & SR$_{0.75}$(\%)                                                   \\ \shline
        \multirow{2}{*}{Center} & -          & 69.1                                                     & 76.5                                                     & 65.1                                                     \\
                                & \checkmark & 68.9\fontsize{9.0pt}{\baselineskip}\selectfont{(-0.2)} & 76.3\fontsize{9.0pt}{\baselineskip}\selectfont{(-0.2)} & 63.5\fontsize{9.0pt}{\baselineskip}\selectfont{(-1.6)} \\ \hline
        \multirow{2}{*}{Corner} & -          & 67.6                                                     & 74.2                                                     & 61.0                                                     \\
                                & \checkmark & 68.7\fontsize{9.0pt}{\baselineskip}\selectfont{(+1.1)} & 76.3\fontsize{9.0pt}{\baselineskip}\selectfont{(+2.9)} & 61.7\fontsize{9.0pt}{\baselineskip}\selectfont{(+0.7)} \\ \hline
        \multirow{2}{*}{Dist.}  & -          & 71.5                                                     & 80.1                                                     & 69.9                                                     \\
                                & \checkmark & 73.6\fontsize{9.0pt}{\baselineskip}\selectfont{(+2.1)} & 83.4\fontsize{9.0pt}{\baselineskip}\selectfont{(+3.3)} & 71.2\fontsize{9.0pt}{\baselineskip}\selectfont{(+1.3)} \\ 
        \shline
    \end{tabular}
    \caption{
        Performance on GOT-10k validation set with Joint Learning(J.L.) integrated in different heads.
    }\label{integration}
\end{table}

%% file: lasot_tknet_table.tex
\begin{table}
    \centering
    \small
    \resizebox{\linewidth}{!}{
        \begin{tabular}{l|cc|cc}
            \shline
                                             & \multicolumn{2}{c|}{TrackingNet} & \multicolumn{2}{c}{LaSOT}                                     \\
            \cline{2-5}
            \rule{0pt}{8pt}
                                             & AUC                              & P$_\mathrm{norm}$         & AUC           & P$_\mathrm{norm}$ \\
            \hline
            JN-384(Ours)                     & \textbf{84.9}                    & \textbf{89.2}             & \underline{69.0} & 78.5              \\
            JN-256(Ours)                     & \underline{84.1}                    & \underline{88.6}             & 68.2          & 78.2              \\
            OSTrack\citeyearpar{ostrack}     & 83.1                             & 88.1                      & \textbf{69.1} & \underline{78.7}     \\
            AiATrack\citeyearpar{aiatrack}   & 82.7                             & 87.8                      & \underline{69.0} & \textbf{79.4}     \\
            MixFormer\citeyearpar{mixformer} & 83.1                             & 87.8                      & 68.1          & 77.2              \\
            KeepTrack\citeyearpar{KeepTrack} & -                                & -                         & 67.1          & 77.2              \\
            STARK\citeyearpar{STARK}         & 82.0                             & 86.9                      & 67.1          & 77.0              \\
            TransT\citeyearpar{transt}       & 81.4                             & 86.7                      & 64.9          & 73.8              \\
            TrDiMP\citeyearpar{TrDiMP}       & 78.4                             & 83.3                      & 63.9          & -                 \\
            Ocean\citeyearpar{Ocean}         & -                                & -                         & 56.0          & 65.1              \\
            ATOM\citeyearpar{Ocean}          & 70.3                             & 77.1                      & 51.5          & 57.6              \\
            \shline
        \end{tabular}
        
    }
    \vspace{-2mm}
    \caption{
        Comparison with the state-of-the-art trackers on TrackingNet and LaSOT.  The best two results are shown in \textbf{bold} and \underline{underline}.}
        \label{table:tknet_lasot}
        \vspace{-4mm}
\end{table}

%% file: vot2020_table.tex
\begin{table*}[!ht]
    \centering
    \small
    \setlength{\tabcolsep}{1.5mm}{

        \begin{tabular}{c|ccccccccc|cc}

            \shline
            \rule{0pt}{8pt}
                       & Ocean   & Atom    & D3S     & DIMP    & STARK   & AiATrack   & RPT\citeyear{RPT}              & MixFormer           & OSTrack & JN-256              & JN-384           \\ \shline
            EAO        & 0.430 & 0.271 & 0.439 & 0.305 & 0.505 & 0.530 & 0.530          & 0.535             & 0.524 & \underline{0.537} & \textbf{0.544} \\
            Accuracy   & 0.693 & 0.462 & 0.699 & 0.492 & 0.759 & 0.764 & 0.700          & 0.761             & 0.767 & \underline{0.790} & \textbf{0.796} \\
            Robutsness & 0.754 & 0.734 & 0.760 & 0.745 & 0.817 & 0.827 & \textbf{0.869} & \underline{0.854} & 0.816 & 0.822             & 0.849          \\ \shline
        \end{tabular}}
    \caption{
        Comparison with the state-of-the-art trackers on VOT2020.  The best two results are shown in \textbf{bold} and \underline{underline}.
    }\label{table:vot2020}
\end{table*}

%% file: inception_table.tex
\begin{table}[]
    \centering
    \begin{tabular}{c|c|c|c}
    \shline
    Inception & Encoding & \#Tokens & AO\%   \\ \hline
    none$^*$      & -        & -           & 73.6 \\ 
    add       & Conv     & -           & 73.0\fontsize{9.0pt}{\baselineskip}\selectfont${(-0.6)}$ \\
    concat    & MLP      & $4$           & 74.2\fontsize{9.0pt}{\baselineskip}\selectfont${(+0.6)}$ \\
    concat    & MLP      & $1$           & 74.3\fontsize{9.0pt}{\baselineskip}\selectfont${(+0.7)}$ \\
    concat    & Sin-Cos  & $1$           & 73.9\fontsize{9.0pt}{\baselineskip}\selectfont${(+0.3)}$ \\ 
    \shline
\end{tabular}
\caption{
    Different ways of inception brings different effect on performance. $*$ indicates the baseline with Joint Learning under positve ratio $\rho=0.7$.
}
\label{table:inception}
\end{table}

%% file: supplementary.tex
\newcommand{\beginappendix}{%
        \setcounter{section}{0}
        \renewcommand\thesection{\Alph{section}}
        \setcounter{table}{0}
        \renewcommand{\thetable}{A\arabic{table}}%
        \setcounter{figure}{0}
        \renewcommand{\thefigure}{A\arabic{figure}}%
     }

\title{Towards Efficient Training with Negative Samples in Visual Tracking}




\beginappendix
\section*{Appendix}

\section{Details in Encoder}
The encoder is identical to the one used in Vision Transformer (ViT).
We start by dividing a given image pair, comprising a template and a search region, into smaller image patches. Specifically, we denote the template image patch as $z \in \mathbb{R}^{3\times H_z \times W_z}$ and the search region patch as $x \in \mathbb{R}^{3\times H_x \times W_x}$.
These patches are then transformed into tokens, $\boldsymbol{H}_{z} \in \mathbb{R}^{D}$ and $\boldsymbol{H}_{x} \in \mathbb{R}^{D}$, using a linear projection layer. We also introduce a novel target-indicating token(TIT), $\boldsymbol{H}_{\texttt{i}} \in \mathbb{R}^{D}$, which aims to enhance the tracking precision by providing location-specific information.
We feed the concatenated tokens, $[\boldsymbol{H}_{\texttt{i}} ; \boldsymbol{H}_{z} ; \boldsymbol{H}_{x}]$, into a transformer encoder similar to ViT. This step facilitates simultaneous feature extraction and relation modeling, thus called one-stage.  One may notice that the information carried by TIT is only related to the template.  However, we found no difference on performance between (1) allowing the TIT to attend to all tokens from both the template and search region and (2) permitting the TIT to attend solely to tokens from the template.  For simlicity, we follow the setting in (1). The final tokens from the search region are selected and flattened to the 2D features before being input to the head for classification and localization.

\section{$[\texttt{cx,cy,w,h}]$ v.s. $[\texttt{w,h}]$ in Target-Indicating Token}

As the template is a center-cropped image of the target to be tracked, the bounding box of the target is always centered.  Thus if we formulate the bouding box as center point and the size $[\texttt{cx,cy,w,h}]$, then the center part $[\texttt{cx,cy}]$ is always $[\texttt{0,0}]$.  Therefore, here comes a question: is the first two number $[\texttt{cx,cy}]$ necessary? We conduct extra experiments on the formulation of the source to the Target-Indicating Token(TIT). As shown in Fig. \ref{table:inception_discuss}, in cotrast to our intuition, $[\texttt{cx,cy,w,h}]$ is more benefitial to the performance than merely $[\texttt{w,h}]$.  The potential reason is even if the target is always located in the center, the model has to learn this prior knowledge by itself.  As the powerful multi-modality of the Transformer, providing more information related to the task helps the model perform better.

\section{ Discretizing Resolution of Distance}
\begin{figure}[t]
       \includegraphics[width=1.0\linewidth]{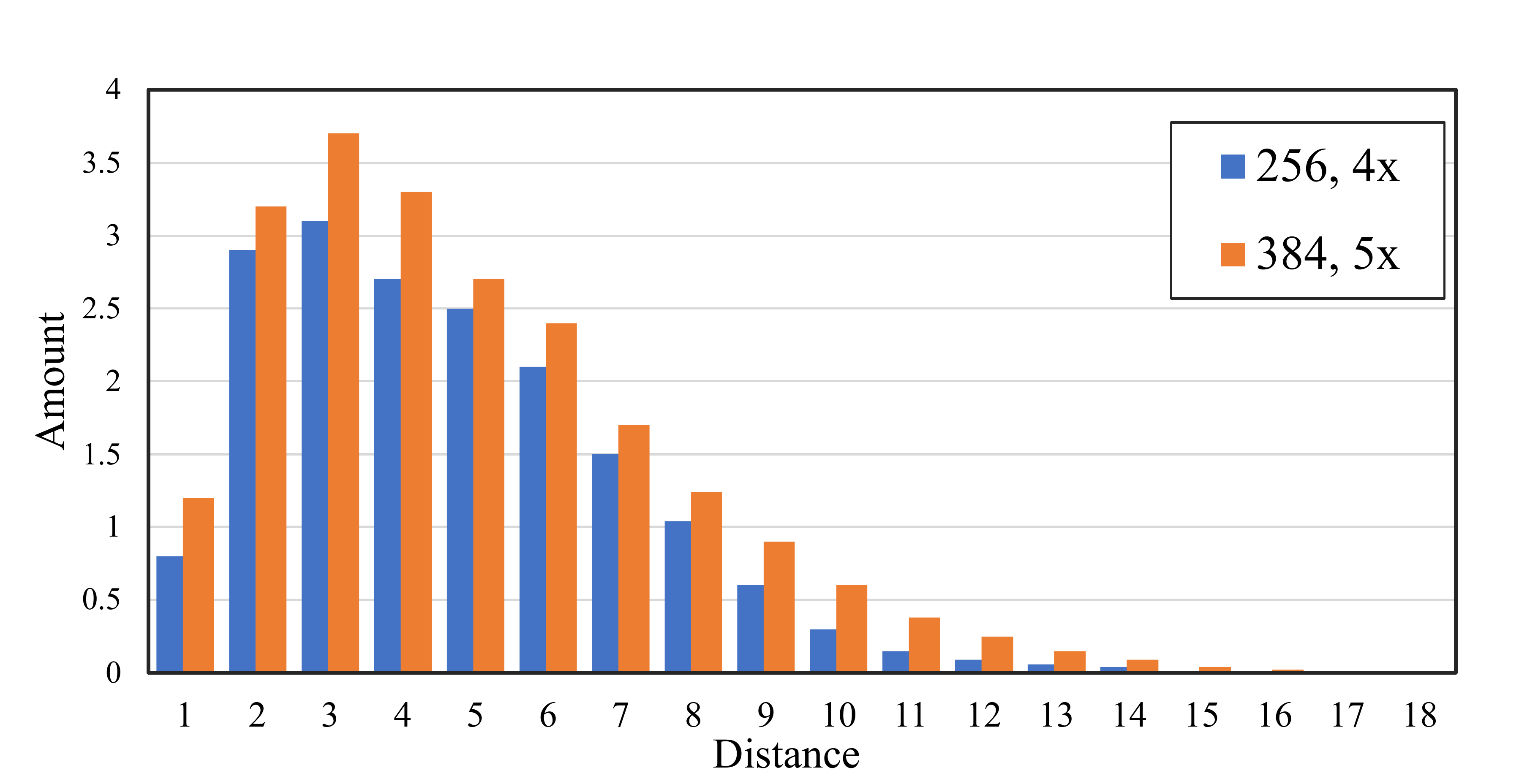}
    
       \caption{The histogram of bounding box regression targets of points inside the box over all training samples on train split of GOT-10k. Since we crop the image to different search region sizes, the distributions of distance varies.  Following the convention, the search region is 4x of target bounding box in 256$\times$256 input setting, while it is 5x in 384$\times$384 setting.}
    \label{distance}
    \end{figure}

Different from object detection, there is only one target to regress in one image in single object tracking(SOT).  The analysis of the distribution of distance over all training samples in GFL\cite{GFL} is under the sample assignment from ATSS\cite{ATSS}.  In SOT, the training fashion is to crop an image around the target with $N^2$ times area as the search region.  With different resolutions for network inputs and various jittering of center position and scale, the statistical distribution of distance may differ.  Here we follow the region crop settings from OSTrack\cite{ostrack}, making a histogram of bounding box regression targets of points inside the box over all training samples on train split of GOT-10k\cite{GOT-10k}.  As shown in Fig.~\ref{distance}, using $n=16$ directly from GFL fits both 256$\times$256 with $4^2\times$ cropped region area and 384$\times$384 with $5^2\times$ cropped region area in the tracking pipeline as well.  Therfore, we keep the  discretizing resolution $n = 16$ in Eq. (1).

\section{Difference with UAST}
UAST\cite{UAST} is the first tracker who adopt the distribution-based head design from GFL\cite{GFL} and GFLv2\cite{GFLv2} in object tracking.  UAST mainly contributes to improve the original head design and make it more appropriate for tracking and bring in the concept of localization quality into the community from object detection.
However, in our method, we simply borrow the head design from GFL.  We do not claim it  as our contribution. Benefit from the distribution-based head for its ability of handling the absence of targets, it helps us train the model with both positive and negative sampels simultaneously.  Our goal is to expolre a more efficient and effective training strategy for visual object tracking.
\section{Limitation}
\begin{table}[]
    \centering
    \begin{tabular}{c|c|c}
        \hline
        Inception & Source                 & AO\%                                                       \\ \hline
        none      & -                      & 73.6\\
        concat    & $[\texttt{w,h}]$       & 73.9\fontsize{9.0pt}{\baselineskip}{(+0.3)} \\
        concat    & $[\texttt{cx,cy,w,h}]$ & 74.3\fontsize{9.0pt}{\baselineskip}{(+0.7)} \\ \hline
    \end{tabular}
    \caption{
        Different ways of inception brings different effect on performance.
    }
    \label{table:inception_discuss}

\end{table}

\begin{table}
    \centering
    \begin{tabular}{c|c|ccc}
        \hline
        Model   & \#Epochs         & AUC(\%) & P(\%)  & P$_\mathrm{norm}$(\%) \\
        \hline
        OSTrack & 100 & 62.2  & 65.9 & 70.1                \\
        JN      & 60 & 63.2  & 66.9 & 71.2                \\
        \hline
    \end{tabular}
    \caption{
        Performance on LaSOT benchmark using the models trained with GOT-10k train split only.
    }
    \label{tabel:lasot_disscus}
\end{table}
\begin{figure}[t!]
       \includegraphics[width=1.0\linewidth]{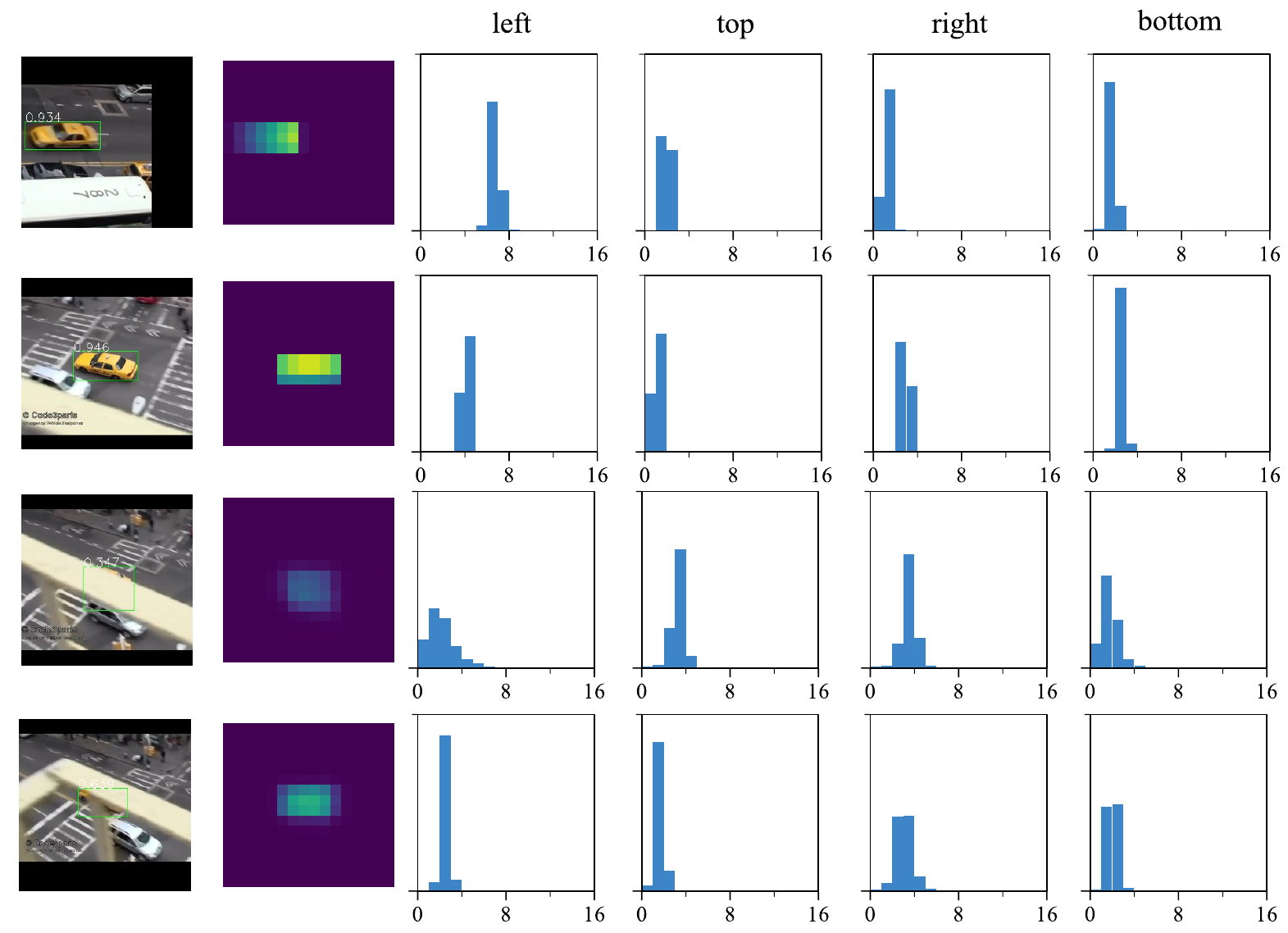}
    
       \caption{The visualization of attention map and the final classification map of each layer.}
    \label{taxi}
\end{figure}
\begin{figure}[t!]
       \includegraphics[width=1.0\linewidth]{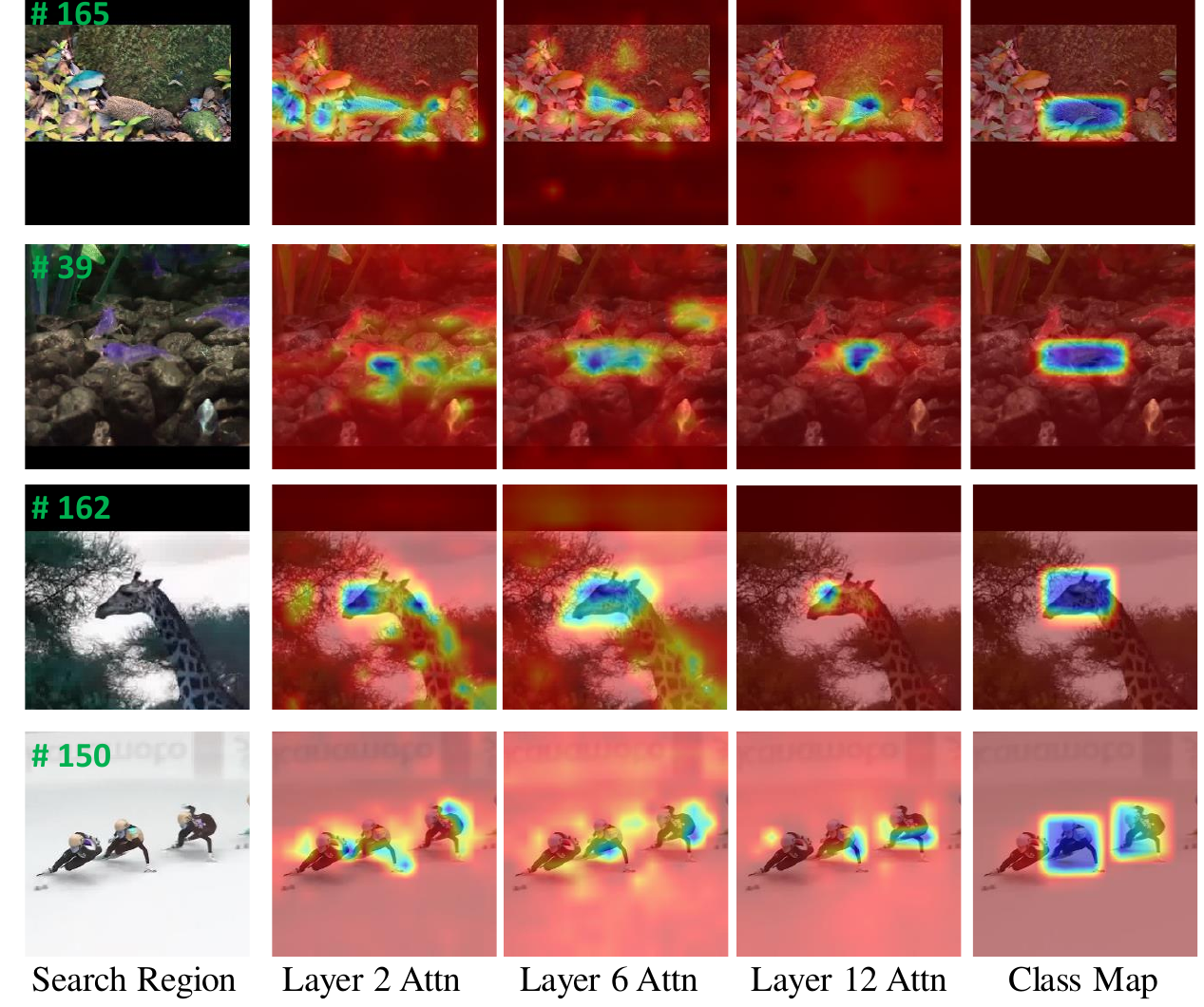}
    
       \caption{The visualization of the distribution of distance from the point with highest classification score.  X-axis in each distribution diagram denotes for the non-normalized distance to the starting point.}
    \label{attn}
\end{figure}

\begin{figure*}[]
    \centering
    \includegraphics[width=0.85\textwidth]{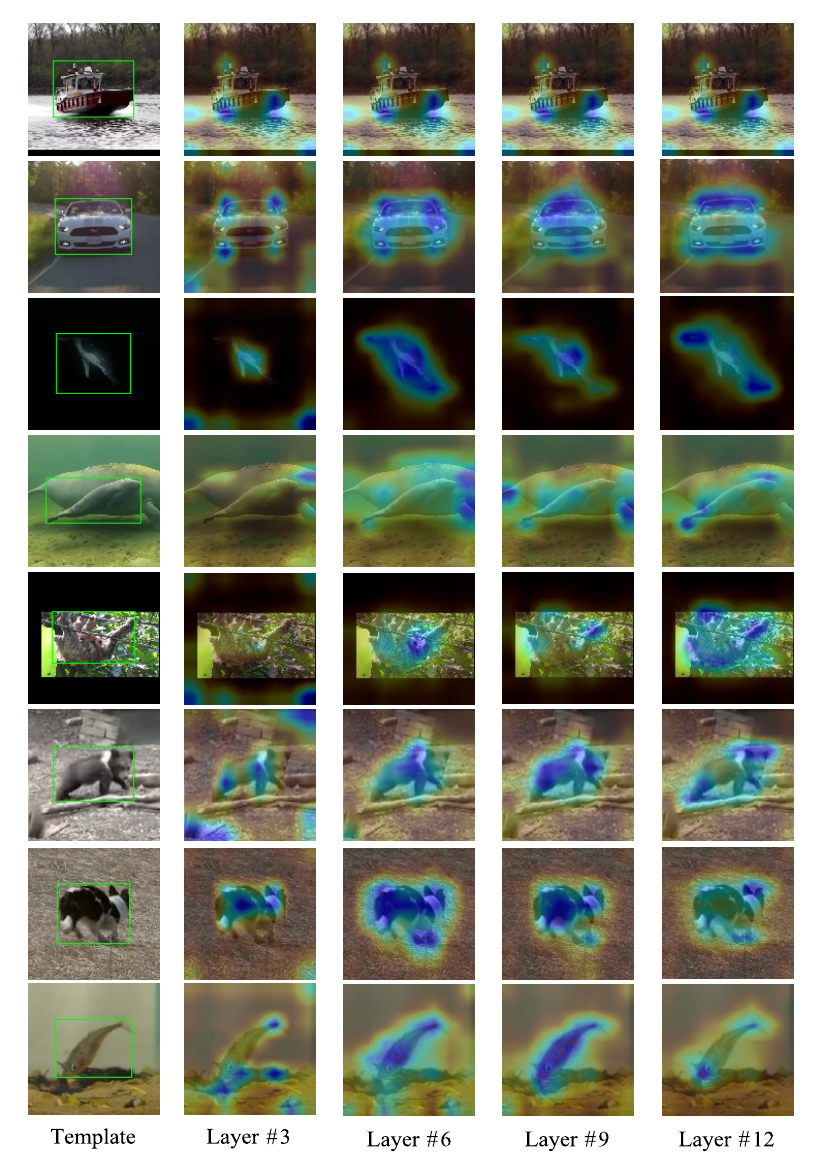} 
    \caption{Visualization of attention map (of TIT attending to the tokens from the template) in the encoder.}
    \label{attn_map}
\end{figure*}
Our proposed tracker achieve not so much competitive performance on LaSOT. Despite the number of sequences being less than that in GOT-10k, TrackingNet and VOT chanllange, the average length of each sequence in LaSOT is much longer, making it a long-term tracking benchmark.  Long-term tracking induces more distributors and  deformation of appearance in temporal, thus requiring the tracker to predict a more accurate classification map.  As shown in Fig.~\ref{attn}, due to the multi-sample assignment in the training, the classification branch is encouraged to predict a box-like class map.  Compared with a single-point class map e.g. center class map, the most confident point predicted by the model may locate at the corner of the target bounding box, which brings more ambiguity.  However, the approach we proposed of adding negative samples is to encourage the model to have more ability of generalization instead.  To verify the generalization of our method, we test the GOT-10k trained models on LaSOT benchmark.  As shown in Tab.~\ref{tabel:lasot_disscus}, our model JN-256 surpasses the OSTrack-256 by 1.0\% AUC, which again shows the generalization of our model when facing the unseen scenarios.

\section{More Visualization}
We first provide more visualization of the distribution-based head in Fig.~\ref{taxi}.  The ouput distribution of the head is sharp when it is certain about the boundary of the target. Due to occlusion, the boundaries of objects are not clear enough, thus the model predicts uncertain distribution.  We notice that most of the output distance is distributed on the samll range, e.g. all the output probability is lower then $0.05$ when $x \in (8,16)$ in Fig.~\ref{taxi}.  We keep this setting because we supervise all the point inside the G.T. bouding box for localization during traing.   The distance from the starting point to the boundary in four directions may various if the starting point located in the corner of the target.

In Fig.~\ref{attn}, we provide more visualization results for attention weights of the search region corresponding to the center part of the template (which can be seen as the target) . The results show that the model attends to the foreground objects at an early stage.

In Fig.~\ref{attn_map}, we visualize more attention map from the TIT to other image tokens in the template by averaging attention weights across heads, reshaping to 2D, and upscaling to the image's resolution.
